\documentclass[journal]{IEEEtran}

\usepackage{times}

\usepackage{url}

\usepackage[breaklinks]{hyperref}
\usepackage{graphicx}
\usepackage{caption}
\usepackage{color,soul}
\usepackage{siunitx}
\usepackage{xcolor}
\usepackage[inline]{enumitem}
\usepackage{amsmath}

\usepackage[noadjust]{cite} 

\begin{document} 

\title{A Flexible Exoskeleton for Collision Resilience} 

\author{Ricardo de Azambuja$^{1,2}$
        Hassan Fouad$^{2}$,
        and~Giovanni Beltrame$^{2}$
\thanks{$^{1}$ University of Edinburgh.}
\thanks{$^{2}$ Polytechnique Montr\'eal.}
\thanks{Contact: ricardo.azambuja@gmail.com}
}

\maketitle 

\begin{abstract}
  With inspiration from arthropods' exoskeletons, we designed a
  simple, easily manufactured, semi-rigid structure with flexible
  joints that can passively damp impact energy. This
  exoskeleton fuses the protective shell to the main robot
  structure, thereby minimizing its loss in payload capacity. Our
  design is simple to build and customize using cheap components and
  consumer-grade 3D printers. Our results show we can build a
  sub-\SI{250}{\gram}, autonomous quadcopter with visual navigation
  that can survive multiple collisions, shows a five-fold increase in
  the passive energy absorption, that is also suitable for automated
  battery swapping, and with enough computing power to run deep neural
  network models. This structure makes for an ideal platform for
  high-risk activities (such as flying in a cluttered environment or
  reinforcement learning training) without damage to the hardware or
  the environment.
\end{abstract}

\vspace*{-0.4cm}

\section{Introduction}
When collisions are hard to avoid, the main strategy for Uncrewed Aerial Vehicle (UAV) designs 
have been the use of protective structures mostly made of materials that are hard themselves. They usually vary 
from fixed structures~\cite{khedekar2019contact}, to rotating ones~\cite{briod2014collision, salaan2019development}.


Instead of only a cage, a frame that is capable of absorbing impact energy is another option for collision resilient
drones. Despite that, previous works~\cite{shu2019quadrotor, mintchev2018bioinspired} 
make it very difficult for the UAV to instantly recover because they
usually lead to an inevitable fall to the ground as they automatically
fold or disconnect the motors. Yet, one significant advantage of not using any special structure to
keep propellers from touching obstacles is the increase in
payload capability.
Nevertheless, unprotected propellers don't allow UAVs to physically
interact with the external world.

\begin{figure}[!t]
	\centering
	\includegraphics[width=0.90\linewidth]{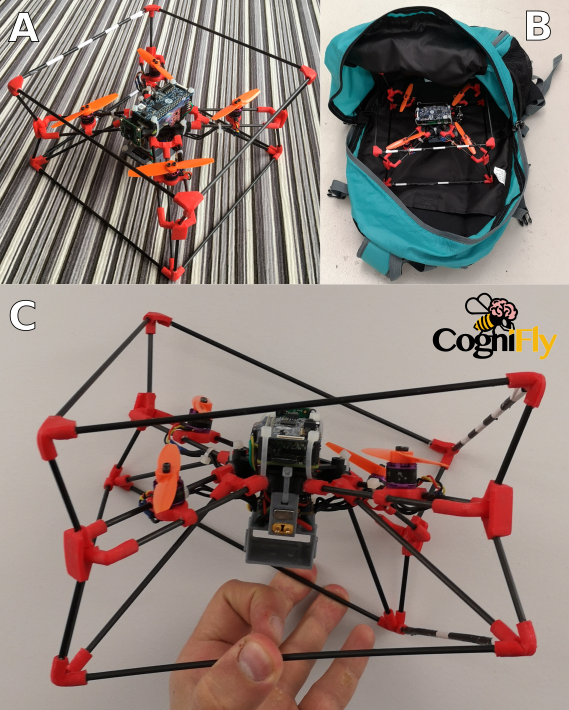}
	\caption{CogniFly (A, B and C) is a small (210x210x120mm), 
	      under-\SI{250}{\gram}, open source, collision
          resilient UAV, capable of autonomous flight (no GPS) and
          complex model inference from on-board equipment. 
		  It mixes soft (red) and rigid (black/gray) 
		  bending to absorb impact energy~(C).}
	\label{fig:cognifly}
\end{figure}

Arthropods are an interesting source of inspiration for innovative UAV designs.
Cockroaches can transition faster by hitting walls~\cite{jayaram2018transition}. 
All arthropods have an \emph{exoskeleton}, and it has a
dual-purpose: \emph{it works as support and protective structure}. 
Nevertheless, it is not necessarily fully rigid, allowing bending
and compression. For an UAV, such a
mixed structure helps increasing the maximum payload and could afford
a simpler control system that allows the UAV to physically interact
with the environment without damage.

Many regulatory agencies use the \SI{250}{\gram} as the threshold for the need of special licenses. However, only a few collision resilient designs keep the weight below \SI{250}{\gram}~\cite{mulgaonkar2017robust,
  sareh2018rotorigami,  mintchev2018bioinspired}. It is also rare to find research drones under \SI{250}{\gram} offering
the extra power to run computationally complex software on-board
(e.g. object detection using neural network
models~\cite{palossi201964, li2020visual}).

The use of truly exoskeleton-like structure mixing rigid and soft materials in a
sub-\SI{250}{\gram} quadcopter has not been studied or presented yet. Some works like~\cite{mintchev2018bioinspired} 
explored flexible frames, but propellers are mostly exposed,
you can't manufacture it without specialized materials or
tools, and they lack the ability to fly autonomously without an external
system.

Here we present the CogniFly (Figure~\ref{fig:cognifly}). 
It is inpired by arthropods, resulting in an exoskeleton made of carbon fiber rods,
connected by 3D printed soft joints. It is designed 
to protect sensitive components and passively damp impacts. 
CogniFly's design depends only on readily available carbon fiber rods and 3D
printers. Also, it can run reasonably complex algorithms and custom deep neural networks
models despite of its relatively small size and weight, which opens the doors for
several potential applications like agriculture, subterranean
exploration, drone swarming and many others.


\section{Materials and methods}

\subsection{Structural design and onboard payload composition}
\begin{figure}[!htb]
	\centering
	\includegraphics[width=0.90\linewidth]{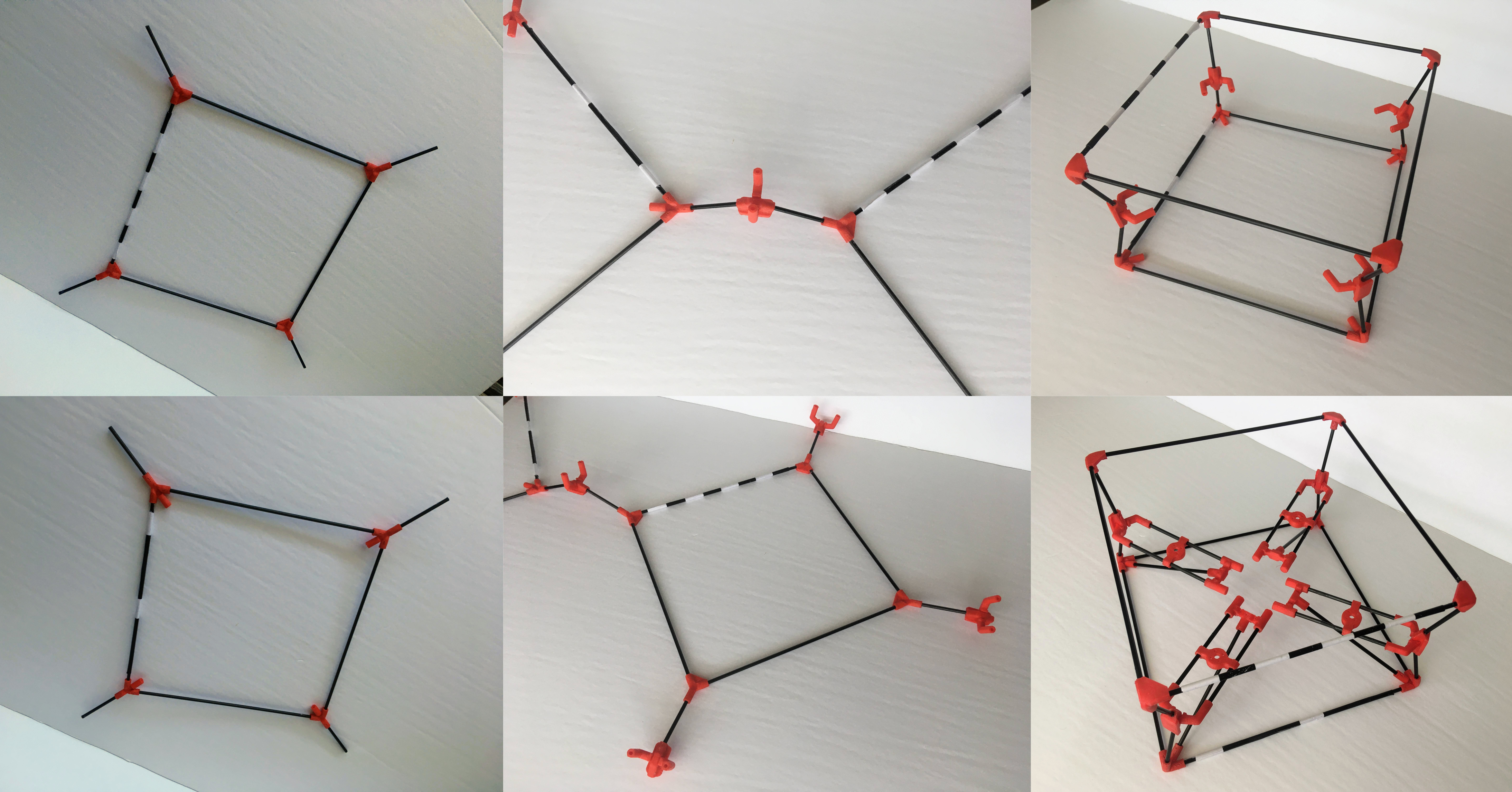}
	\caption{Steps (left to right, top to bottom) to assemble the outer frame by folding
	the flexible joints. After the last step, it is necessary 
	to add the rigid central 3D printed parts to have a full, ready-to-fly frame.}
	\label{fig:folding_as_you_assemble}
\end{figure}

The CogniFly takes inspiration from arthropods’ exoskeletons in that its protective cage 
is a part of the overall structure. 
The protective role of the exoskeleton calls for the ability to damp
and distribute impact energy, which leads to the use of 3D printed flexible 
joints (TPU 95A) to connect the carbon fiber rods (Figure~\ref{fig:folding_as_you_assemble}). This material enables the drone
to be generally flexible, as shown in
Figure~\ref{fig:cognifly}-C, which helps dissipating impact
energy, while keeping the shape of its central part, allowing
operation with a regular flight controller. 

CogniFly uses a Raspberry Pi Zero W with a camera module v2,
and it runs complex models using Google's AIY Vision
Bonnet (also supporting the Coral USB Accelerator). It features an Optical Flow and Time-of-Flight sensors,
thus CogniFly can operate autonomously. Its battery holder and sleeve are designed to
enable easy and quick automated extraction and insertion of batteries\footnote{More details available at thecognifly.github.io}.

\subsection{Free fall crash landing experiments}
\begin{figure}[!htb]
	\centering
	\includegraphics[width=0.90\linewidth]{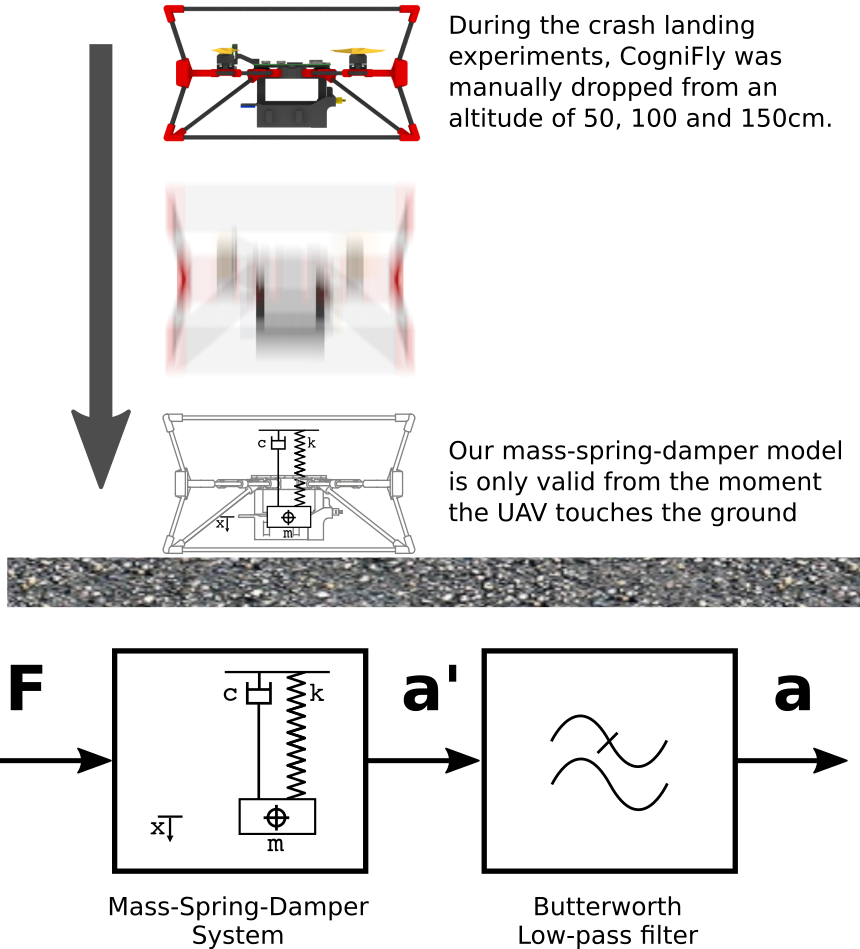}
	\caption{Free-fall experiments~(top) and
		block diagram~(bottom). Three drop altitudes were used: $50$, $100$ and \SI{150}{\cm})}
	\label{fig:experiment}
\end{figure}

We reckon a vertical free fall to be a critical scenario as we
consider payload contact with hard exterior objects, like the ground,
has the highest potential of causing damage.  Moreover, the battery
is located at the bottom part of the drone, and it should not be subjected to
extreme mechanical loads. The main criterion we adopt in this regard is the
maximum absolute acceleration of the payload in case of a vertical
free fall impact (i.e. crash landing), depicted in
Figure~\ref{fig:experiment}.
To test our exoskeleton design, we perform a series of free fall crash landing experiments and record the absolute
acceleration using the ADXL377 (3-Axis, $\pm$200G, 500Hz) to
assess the UAV's ability to absorb impact energy by comparing between the values recorded from the flexible CogniFly's frame (\SI{241}{\gram})
and only the rigid central part of the frame made entirely of ABS~(\SI{239}{\gram}).

\subsection{Mass-spring-damper model}
The impact absorbing aspect of the CogniFly for the free fall experiments is modelled as a
mass-spring-damper system ($m\ddot{x} + c\dot{x} + kx = F$) 
where $m>0$ is the mass, $c>0$ is the equivalent damping
coefficient of the damper and $k>0$ is the equivalent stiffness of the
spring. Moreover, we augment the previous
model with a first order butterworth low pass filter ($fc=500Hz$) 
to model the sampling latency.

A loss function is created to estimate the equivalent damping~($c$)
using the Mean Square Error (MSE) between our model
(Figure~\ref{fig:experiment}, bottom) and the collected acceleration peak data for all experiments. 
The value of the equivalent stiffness is 
obtained through a unidimensional static deformation test.

The mass-spring-damper model allows the estimation of the energy that goes 
into different parts of the system during impacts~(Figure~\ref{fig:experiment}). 
As a simplification, our model can't take into account the collision between
the payload and the ground and it is only valid until $x\leq16mm$. 
Therefore, in these situations we calculate the energy dissipated by the damper 
considering the difference between the initial kinetic energy and the kinetic
energy when~$x=16mm$.

\section{Results} 
In the following, we present results that show CogniFly's ability to absorb impacts and how 
our mass-spring-damper model helps understanding the overall structural behaviour.

\begin{figure}[!htb]
	\centering
	\includegraphics[width=0.90\linewidth]{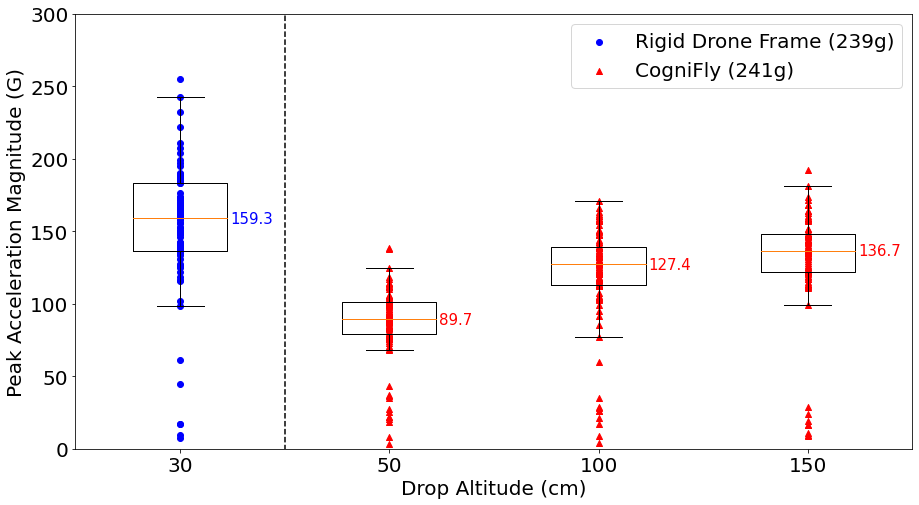}
	\caption{Boxplot and scatter plot of all collected
          experimental crash landing peak acceleration data for CogniFly (right) 
		  and a rigid structure~(left).}
	\label{fig:crash_landing_results}
\end{figure}

We carry out the free fall experiments for the CogniFly at three different
altitudes: \SIlist{50;100;150}{\cm}. To avoid irreparable damage to 
the rigid structure used in the comparisons, we limit its free fall to \SI{30}{\cm}. 
Figure~\ref{fig:crash_landing_results} shows the median of the absolute acceleration peak values for the rigid frame
falling from \SI{30}{\cm} is higher than that of the CogniFly falling
from \SI{150}{\cm}, a five-fold~($150/30$) improvement. 

\begin{figure}[!htb]
	\centering
	\includegraphics[width=0.90\linewidth]{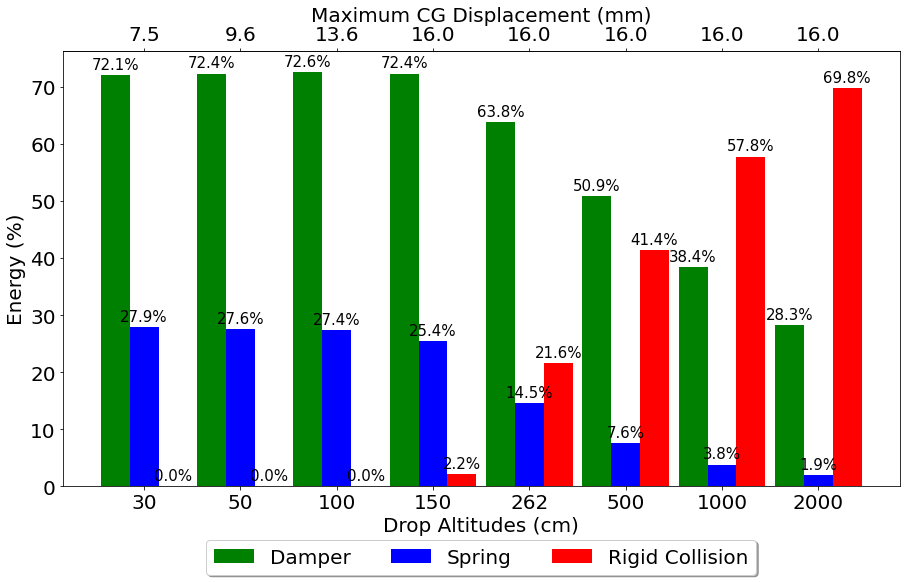}
	\caption{Energy distribution according to the
          Mass-Spring-Damper model (Figure~\ref{fig:experiment},
          $m=0.241$, $c=46$ and $k=7040$). In this plot, we consider a
          collision when the CG displacement reaches 16mm.}
	\label{fig:model_energy_bars}
\end{figure}

Using the data collected from the experiments (Figure~\ref{fig:crash_landing_results}), 
the minimization of the loss and the static measurements resulted in a mass-spring-damper model with 
coefficients $c=46$ and $k=7040$. Using this model, we predict the distribution
of energy stored and dissipated. The results (Figure~\ref{fig:model_energy_bars}) show a graceful degradation where the energy that will
eventually cause a failure (collision) doesn't increases in
steeply and even at 20m (2000cm) more than $30\%$ of the
initial potential energy will end up being dissipated
(damper) and stored (spring) by the frame according to our model.

For altitudes below \SI{150}{\cm}, Figure~\ref{fig:model_energy_bars}
shows that the total kinetic energy is divided between the damper and
the spring, while for higher altitudes the amount of energy that goes
into rigid collision increases with altitude. Such collision energy
can give an indication of how strong the impact between payload and
ground is, and this can help deciding how far the operational altitude
of the drone can be pushed, had there been a need to go beyond a safe
altitude.

As a final test, CogniFly was dropped multiple times from 262cm (the maximum
altitude our experimental setup allowed us, literally our ceiling).
Although the drone reached a speed at impact of approximately \SI{7}{\meter\per\second}, 
it didn't sufferred	any damage. Compared to some of the latest works on collision resilience UAVs with 
equivalent size and weight \cite{sareh2018rotorigami, shu2019quadrotor, 
 zha2020collision}, CogniFly reached a higher speed 
at impact without damages.

\section{Discussion and Conclusions}
In this paper, we introduce a new paradigm for collision resilient
UAV design inspired by the flexible exoskeleton of arthropods, fusing
the protective cage and the main frame in one semi-rigid structure
with flexible joints that can withstand high-velocity impacts. Our
UAV~(Figure~\ref{fig:cognifly}), weighs under \SI{250}{\gram} and blends
rigid and soft materials giving the final structure the ability to
absorb and dissipate impact energy, while still being sufficiently
stiff for agile flight. Thanks to its exoskeleton, it is possible to
save precious payload capability when compared to a traditional
protective cage design.

CogniFly survived multiple collisions at speeds up to
\SI{7}{\meter\per\second} while carrying enough computing power to run
deep neural network models. Throughout a series of simple crash
landing experiments (Figure~\ref{fig:experiment}), we show CogniFly
withstands up to a five fold increase in the maximum collision energy
when compared to a rigid system (3D printed on ABS) of similar
weight. Moreover, we employ the experimental data to create a lumped
mass-spring-damper model that allows us to extrapolate the results to
untested cases while the calculated damping and stiffness can be used
in to better understand the role of different materials or
configurations.

We designed CogniFly from the ground up for easy manufacturing and it 
can be built using a very small consumer-grade 3D printer, 
in addition to inexpensive off-the-shelf parts. The design of the drone itself 
was restricted by maximum weight (below \SI{250}{\gram}) and size (able to fit in a backpack
without any disassembly, Figure~\ref{fig:cognifly}). 

Finally, as this is an open source project, all 3D files, software and
data will be freely available under \href{https://github.com/thecognifly}{The CogniFly Project}.

\section*{Acknowledgments}
We would like to thank the financial support received from
\href{https://ivado.ca}{IVADO} (postdoctoral scholarship 2019/2020)
and the productive discussions and help received from current and past
students and interns from \href{https://mistlab.ca}{MISTLab}.

\bibliography{CogniFly}
\bibliographystyle{unsrt}

\end{document}